\documentclass{article}
\usepackage{spconf}
\usepackage{ifpdf}
\usepackage{tcolorbox}
\tcbuselibrary{theorems}
\usepackage{amsmath}
\usepackage{amsfonts}
\usepackage{amssymb}
\usepackage{subcaption}
\usepackage{hyperref}

\usepackage{cleveref}

\usepackage{appendix}
\newcommand{\red}[1]{ {\color{red}#1} }

\definecolor{my-blue}{cmyk}{0.15, 0.0, 0.0, 0.9, 1.00}
\definecolor{my-gray}{cmyk}{0.05, 0.0, 0.0, 0.05, 1.00}
\usepackage{booktabs}
\usepackage{algorithm}
\usepackage{algpseudocode}
\usepackage{algcompatible}
\usepackage{enumitem}
\newlength{\maxwidth}
\newcommand{\algalign}[2]
{\makebox[\maxwidth][r]{$#1{}$}${}#2$}
\newcommand\blfootnote[1]{%
  \begingroup
  \renewcommand\thefootnote{}\footnote{#1}%
  \addtocounter{footnote}{-1}%
  \endgroup
}

\usepackage{spconf,amsmath,graphicx}
\usepackage{tcolorbox}
\usepackage{lipsum} 
\usepackage{multirow}
\def\x{\mathbf{x}}

\DeclareMathOperator*{\minimize}{minimize}

\title{Neural networks with quantization constraints.}
\name{Ignacio Hounie*, Juan Elenter* and Alejandro Ribeiro. \thanks{Supported by NSF-Simons MoDL, Award 2031985, NSF AI Institutes program, Award 2112665, and NSF HDR TRipods Award 1934960. * Equal contribution.}}

\address{University of Pennsylvania}

\begin{document}
\ninept
\maketitle
\begin{abstract}
Enabling low precision implementations of deep learning models, without considerable performance degradation, is necessary in resource and latency constrained settings. Moreover, exploiting the differences in sensitivity to quantization across layers can allow mixed precision implementations to achieve a considerably better computation performance trade-off. However, backpropagating through the quantization operation requires introducing gradient approximations, and choosing which layers to quantize is challenging for modern architectures due to the large search space. In this work, we present a constrained learning approach to quantization aware training. We formulate low precision supervised learning as a constrained optimization problem, and show that despite its non-convexity, the resulting  problem is strongly dual and does away with gradient estimations. Furthermore, we show that dual variables indicate the sensitivity of the objective with respect to constraint perturbations. We demonstrate that the proposed approach exhibits competitive performance in image classification tasks, and leverage the sensitivity result to apply layer selective quantization based on the value of dual variables, leading to considerable performance improvements.
\end{abstract}
\begin{keywords}
Quantization Aware Training, Constrained Optimization, Deep Learning, Duality
\end{keywords}
\section{Introduction}\label{sec:intro}

Despite the success and ubiquity of deep learning models for machine learning tasks, their size and computational requirements can be prohibitive for their deployment on resource or latency constrained settings. Furthermore, as the deployment of these solutions scales, their high energy consumption on inference can pose a challenge to sustainability~\cite{energy-dl-inference}. To tackle this, prior work has proposed leveraging fixed point, low precision hardware implementations. For instance, 8 bit fixed point representations have been shown to reduce model size, bandwidth and increase throughput without significantly compromising performance \cite{Integer-QF-sensitivity}.

However, training models using only low precision implementations can be hard due to the highly non-smooth optimization landscape induced by coarse discretisations~\cite{sharpness-quant}. To overcome this, \emph{model quantization} techniques~\cite{whitepaper-quant} allow to map models from high precision to low precision representations, enabling the use of high precision models/operations during training.

Solely optimising the model in high precision and then performing \emph{post training quantization}~\cite{PTQ-OG} does not account for the performance degradation of high precision solutions after being quantised. Therefore, a plethora of \emph{quantization aware training} methods~\cite{survey-arxiv-22} have been proposed, aiming to exploit both high and low precision representations during optimisation. Nonetheless, using stochastic gradient descent methods is challenging because the quantisation operation does not have meaningful pointwise gradients, and thus several approximation methods have been proposed~\cite{steBengio, Gong2019DifferentiableSQ, cos-reg, non-diff, defossez2021differentiable}. 

In spite of their effectiveness, quantization aware training methods still fail to match the performance of full precision models in low bitwidth settings~\cite{whitepaper-quant}. A myriad of machine learning models consist of the composition of simpler functions, such as layers or blocks in neural networks. Because model performance is not equally sensitive to quantization errors in different layers, mixed precision implementations can provide a better trade-off between computation and performance~\cite{Integer-QF-sensitivity}. For instance, using higher precision for the first and last layers is a common practice in neural network compression literature and often leads to considerable improvements in performance~\cite{BengioBinarizedNN,XNORNetIC,dorefa}. However, analysing the sensitivity to quantization errors after training is limited, because it does not account for the impact of quantization on training dynamics. Although several approaches to determine the optimal bit-width for each layer have been proposed~\cite{survey-arxiv-22}, they often involve computationally intensive search phases.

In this work, instead of trying to optimize low precision performance, we frame quantization aware training as learning a high precision model that is robust to using low precision implementations. To do so, we formulate a constrained learning problem that imposes proximity between the model and its quantized counterpart. Despite being non-convex, the resulting problem has zero duality gap. This enables the use of primal-dual methods, and removes the need to estimate the gradient of the quantization operation. Moreover, imposing layerwise constraints and leveraging strong duality, we show that optimal dual variables indicate the sensitivity of the objective with respect to the proximity requirements at a specific layer. We demonstrate the benefits of our method in CIFAR datasets, using popular quantization techniques.

\section{Problem Formulation}\label{sec:prob-form}

\subsection{Low precision supervised learning}
As in the standard supervised learning setting, let $\x \in \mathcal{X} \subseteq \mathbb{R}^{d}$ denote a feature vector and $ y \in \mathcal{Y} \subseteq \mathbb{R}$ its associated label or measurement. Let~$\mathfrak{D}$ denote a probability distribution over the data pairs~$(\x, y)$ and~$\ell: \mathcal{Y}\times \mathcal{Y} \to  \mathbb{R}_+$ be a non-negative loss function, e.g., the cross entropy loss. The goal of Statistical Learning is to learn a predictor $f: \mathcal{X}\to\mathcal{Y}$ in some convex functional space $\mathcal{F}$, that minimizes an expected risk, explicitly,
\begin{align}\tag{SRM}\label{SRM}
\operatorname{minimize}_{f \in \mathcal{F}} \quad \mathbb{E}_{(\x, y) \sim \mathfrak{D}}\left[\ell(f(\x), y)\right].
\end{align}
In (\ref{SRM}) the input and output spaces $\mathcal{X}$, $\mathcal{Y}$ can be continuous real valued spaces, and $\mathcal{F}$ is an infinite dimensional functional space. However, since practical implementations rely on finite-precision representations and operations, only piece-wise constant functions can be represented. That is, the input and output space, as well as the operations performed, need to be discretised in order to be implemented in digital hardware. Whereas full precision floating point implementations can provide a good approximation of $f$, the use of coarser discretisations exacerbates the approximation error.

We will thus consider a \emph{finite} space of functions $\mathcal{F}^q$ that admit a low precision representation, and define a quantization operation $q: \; \mathcal{F} \to \mathcal{F}^q$ that maps each function in $f \in \mathcal{F}$ to a function $q(f)=f^q \; \in \mathcal{F}^q$. Then, the supervised learning problem under low precision can be formulated as
\begin{align}\tag{Q-SRM}\label{Q-SRM}
\minimize_{f \in \mathcal{F}} \quad \mathbb{E}_{(\x, y) \sim \mathfrak{D}}\left[\ell(f^q(\x), y)\right].
\end{align}
One of the challenges of low precision supervised learning~(\ref{Q-SRM}) is that, since the quantization operator $q$ is a piecewise constant function (it maps $\mathcal{F}$ to the finite set $\mathcal{F}^q$), 
the pointwise gradients\footnote{We are referring to the functional (Fréchet) derivative of the operator $f^q$ as a function of $f$.} of the loss $\ell$ with respect to $f$ are zero almost everywhere; explicitly,
\begin{align*}
\nabla_f \left[\ell(f^q(\x), y)\right] = \frac{\partial f^q}{\partial f}\; \nabla_{f^q} \left[\ell(f^q(\x), y)\right] = 0 \; \text{a.e. }
\end{align*}
Therefore, to enable gradient-based optimization techniques, gradient approximations such as STE \cite{steBengio} are needed.

\subsection{Quantization Robust Learning}\label{subsec:QRL}
Instead of restricting the space $\mathcal{F}$ to those functions that can be implemented \emph{exactly}, we would like to learn a model in $\mathcal{F}$ that is robust to using low precision implementations. We can formulate this as a proximity requirement between $f$ and its quantised version $f^q$. If proximity is measured in terms of model outputs, the requirement can be written as:
$$
d(f(x), f^q(x)) \leq \epsilon \quad \forall \: x \in \mathcal{X},
$$
where $d : \mathcal{Y}\times \mathcal{Y} \to  \mathbb{R}$ is a distance between outputs (e.g: cross-entropy loss). We may also wish to weight different inputs depending on their probability, in order to reduce the impact of unlikely or pathological cases. We achieve this by averaging over the data distribution:
$$
\mathbb{E}_{\mathfrak{D}} \left[ d(f(x), f^q(x)) \right] \leq \epsilon \: .
$$
This leads to the constrained statistical learning problem:
\vspace{-0.04in}
\begin{align*}
P^{\star} = \: & \min_{f \in \mathcal{F}} \quad  \; \mathbb{E}_{\mathfrak{D}} \left[\ell \left(f(\x), y \right)\right] \\
&\text{s.t. :} \quad  \mathbb{E}_{\mathfrak{D}} \left[ d(f(x), f^q(x)) \right] \leq \epsilon \;. \nonumber
\end{align*}

\subsection{Leveraging compositionality}

Since quantization errors can compound in complex ways across different layers, requiring proximity in the outputs of models as in~(\ref{QRL}) does not guarantee proximity for individual blocks. In addition, the output of the model is not equally sensitive to quantization errors in different layers~\cite{HAWQV2}. This can be leveraged by implementing layers or blocks with different levels of precision~\cite{Integer-QF-sensitivity}. 

Thus, we will explicitly address predictors that can be expressed as the composition of $L$ functions $f_{l}: \mathcal{Z}_l \to \mathcal{Z}_{l+1}$, where $\mathcal{Z}_l \subseteq \mathbb{R}^{{d}_l}$. Namely, $f = f_{L} \;o \ldots o \; f_{1}$, where each function $f_l$ belongs to a hypothesis space $\mathcal{F}_l$. As in the previous section we will consider functions $f_l^q \in \mathcal{F}^q_l$ that admit a low precision implementation, and quantization operators $q_{l}: \; \mathcal{F}_l \to \mathcal{F}^q_l$ that map each function in $f_l \in \mathcal{F}_l$ to a function $q_l(f_l)=f^q_l \; \in \mathcal{F}^q_l$. 

We can then formulate a proximity requirement for each layer $f_l$ and its quantised version $f^q_l$, namely,
$$
\mathbb{E}_{\mathfrak{D}} \left[ d_l(f_l(\mathbf{z}^q_{l-1}), f_l^q(\mathbf{z}^q_{l-1})) \right] \leq \epsilon_l \:.
$$
where $\mathbf{z}^q_{l} = f^q_{l} \; o \;\ldots \;o\; f^q_{1}(\x)$ denotes the output of the $l$-th function of the low precision model for an input $\x$. Since each layer and its quantized version are evaluated on the same activations, this notion of proximity is restricted to the error introduced on a single layer or block, doing away with the need to consider error accumulation.

The constraint level $\epsilon_l$ and distance function $d_l$ can be adapted to the particular layer or block considered, leveraging a priori knowledge of the model's architecture and implementation, enabling the use of selective quantization. This leads to the constrained learning problem:

 
\begin{tcolorbox}
\vspace{-0.15in}
\begin{align*}\tag{QRL}\label{QRL}
P^{\star}_c &= \min_{f \in \mathcal{F}}  \quad  \; \mathbf{E}_{(x, y)\sim \mathfrak{D}} \left[\ell \left(f(\x), y \right)\right], \\
& \text{s.t. :} \:   \mathbb{E}_{\mathfrak{D}} \left[ d_l(f_l(\mathbf{z}^q_{l-1}), f_l^q(\mathbf{z}^q_{l-1})) \right] \leq \epsilon_l,  \: \:  l = 1,..,L-1 \nonumber\\
  & \quad \quad \, \mathbb{E}_{\mathfrak{D}} \left[ d(f(x), f^q(x)) \right] \leq \epsilon_{\text{out}} 
\end{align*}
\end{tcolorbox}

The dual problem associated to \eqref{QRL} is:
\vskip -0.15in
\begin{align*}
\tag{D-QRL}
\label{D-QRL}
D^{\star}_c &=  \max_{\lambda_l, \lambda_{out} \geq 0}  \min_{f \in \mathcal{F}} \: \mathbf{E}_{(x, y)\sim \mathfrak{D}}  \large \{ \ell \left(f(\x), y \right) \\ & +  \sum_{l=1}^{L-1} \lambda_l \left( d_l(f_l(\mathbf{z}^q_{l-1}), f_l^q(\mathbf{z}^q_{l-1})) - \epsilon_l \right) \\ & +
\lambda_{out} \left[ d(f(x), f^q(x)) -  \epsilon_{\text{out}} \right] \large\}
\end{align*}
As opposed to the standard \eqref{Q-SRM} formulation, the gradient of the Lagrangian with respect to $f$ is non-zero, 
removing the need to estimate the gradient of the quantization operator $q$. The dual problem can be interpreted as finding the tightest lower bound on $P^{\star}_c$. In the general case, $D^{\star}_c \leq P^{\star}_c$, which is known as weak duality. Nevertheless, under certain conditions, $D^{\star}_c$ attains $P^{\star}_c$ (strong duality) and we can derive a relation between the solution of \eqref{D-QRL} and the sensitivity of $P^{\star}_c$ with respect to $\epsilon_l$ and $\epsilon_{out}$. We explicit these properties in the following section. 

\subsection{Strong duality and sensitivity of \eqref{QRL}}

Note that $f^q=q(f)$ is a non-convex function of $f$, making \eqref{QRL} a non-convex optimization problem. Nevertheless, strong duality can still be leveraged under the following assumptions:

\begin{enumerate}
 \item[] \textbf{(A1)} For all $l$, there exist $ f_l \circ f_{l-1}^q \cdots \circ f_1^q (\mathbf{x}) \in  \mathcal{F}_l \circ \mathcal{F}^q_{l-1} \cdots \circ \mathcal{F}^q_1$ that is strictly feasible (Slater's).
\item[] \textbf{(A2)} The set $\mathcal{Y}$ is finite.
\item[] \textbf{(A3)} The conditional distribution of $\mathbf{x}|y$ is non-atomic.
\item[] \textbf{(A4)} The closure of the set $\mathcal{F}_l \circ \mathcal{F}^q_{l-1} \cdots \circ \mathcal{F}^q_1$ is decomposable for all $l$.
\end{enumerate} 
\vskip -0.05in
\begin{tcolorbox}
\textbf{Proposition 1:} Under assumptions \textbf{(A1-4)} the problem \eqref{QRL} is strongly dual, that is: $$P^{\star}_c = D^{\star}_c .$$
\end{tcolorbox}\textit{Proof. This is a particular case of  \cite{LuizNonConvex}, Proposition III.2. }\\
Note that $\textbf{(A1)}$ is typically satisifed by hypothesis classes with large capacity. We can extend the analysis to regression settings by removing \textbf{(A2)} and adding a stronger continuity assumption on the losses. $\textbf{(A3)}$ is a mild assumption since it only excludes the existence of inputs that are a determinstic function of their output, which holds for processes involving noisy measurements and problems presenting continuous structural invariances. Finally, $\textbf{(A4)}$ is satisfied by Lebesgue spaces (e.g, $L_2$ or $L_{\infty}$), among others.
\\
Note that $P^{\star}_c$ is a function of the constraint tightness $\epsilon_l$. The sub-differential of $P^{\star}_c(\epsilon_l)$ is defined as:
$$
\partial P^{\star}_c(\epsilon_l) = \{ z \in \mathbb{R^{+}} \, : \,P^{\star}_c(\epsilon') \geq P^{\star}_c(\epsilon_l) +  z(\epsilon'-\epsilon_l) \: \: \text{for all $\epsilon'$}\}
$$
In the case where the sub-differential is a singleton, its only element corresponds to the gradient of $P^{\star}_c$ at $\epsilon_l$. Having the above definition, we state following theorem, which characterizes the variations of $P^{\star}_c$ as a function of the constraint tightness $\epsilon_l$.

\vskip 0.02in

\begin{tcolorbox}
\textbf{Theorem 1:} Let $(\mathbf{\lambda}^{\star}_1,\dots,\mathbf{\lambda}^{\star}_{L-1}, \mathbf{\lambda}^{\star}_{\text{out}}) $ be a solution of \eqref{D-QRL}. Under assumptions \textbf{(A1-4)}, 
\begin{align*}
& -\lambda^{\star}_l \: \in \: \partial P^{\star}_c(\epsilon_l) \quad \forall \: l = 1, \cdots, L-1 \\
& -\lambda^{\star}_{\text{out}} \: \in \: \partial P^{\star}_c(\epsilon_{\text{out}})
\end{align*}

\textit{Proof. See Appendix~\ref{app:sensi}}
\end{tcolorbox} 


This implies that the optimal dual variables indicate the sensitivity of the optimum with respect to the proximity requirements at the layer level. This give users information that could be leveraged a posteriori, for example, to allocate higher precision operations to more sensitive layers. Note that these properties also apply to the problem with one output constraint, since it can be recovered from \eqref{QRL} by setting $d_l(\cdot, \cdot) = 0$.

\section{Empirical Dual Constrained Learning}\label{sec:proposed}

The problem \eqref{QRL} is infinite-dimensional, since it optimizes over the functional space $\mathcal{F}$, and it involves an unknown data distribution $\mathfrak{D}$. To undertake \eqref{QRL}, we replace expectations by sample means over a data set $\{(\boldsymbol{x}_i, y_i) \: : \: i=1,\cdots,N \}$ and introduce a \emph{parameterization} of the hypothesis class $\mathcal{F}$ as $\mathcal{F}_{\theta} = \left\{f_{\boldsymbol{\theta}} \mid \boldsymbol{\theta}_{l} \in \Theta_l \subseteq \mathbb{R}^{p_{l}}, \; l = 1,\ldots L \right\}$, as typically done in statistical learning. These modifications lead to the Empirical Dual Constrained Learning problem we present in this section. \\
Recent duality results in constrained Lerning~\cite{LuizNonConvex}, allow us to approximate the problem \eqref{QRL} by its empirical dual:
\begin{align}\tag{ED-QRL}~\label{ED-QRL}
D^{\star}_{\text{emp}} = & \max_{\mathbf{\lambda} \geq 0 } \min_{\boldsymbol{\theta} \in \Theta}, \hat{L}(\boldsymbol{\theta}, \mathbf{\lambda})
\end{align}
where $\hat{L}$ is the empirical Lagrangian of~\eqref{QRL}:
\begin{align*}
&\hat{L}(  \boldsymbol{\theta},  \mathbf{\lambda}) = \frac{1}{N} \sum_{i=1}^N  \ell \left(f_{\mathbf{\theta}}(\x_i), y_i \right)  \\ &+ \sum_{l=1}^{L-1}\lambda_{l} \left[ \left( \frac{1}{N} \sum_{i=1}^N d_l(f_{\mathbf{\theta}_l}(\mathbf{z_i}^q_{l-1}), f_l^q(\mathbf{z_i}^q_{l-1})) \right) - \epsilon_{l} \right ]\\
   &+ \lambda_{\text{out}}\left[\left( \frac{1}{N} \sum_{i=1}^N d(f_{\mathbf{\theta}}(\mathbf{x}_i), f^q(\mathbf{x}_i)) \right) - \epsilon_{\text{out}} \right].
\end{align*} 

\begin{algorithm}[t]
\caption{\textbf{P}rimal \textbf{D}ual \textbf{Q}uantization-\textbf{A}ware \textbf{T}raining (PDQAT).}
    \label{alg:pd}
    \begin{algorithmic}[1]
    \STATE {\bfseries Input:} Dataset $\{x_i, y_i \}_{i=1,\cdots,N} $, primal learning rate $\eta_p$, dual learning rate $\eta_d$, number of epochs $T_e$, number of batches $T_b$, constraint tightness $\epsilon_1,\cdots,\epsilon_{\text{out}}$.
    \STATE Initialize:
    $\begin{aligned}
    \boldsymbol{\theta}, \lambda_1, \dots, \lambda_{L-1}\leftarrow\mathbf{0} \quad \lambda_{out}\leftarrow\mathbf{1}
    \end{aligned}$ \\
    \FOR{$\text{epoch } =1, \ldots, T_e \: $ } \\
    \vskip 0.05in
        \hspace*{2mm} \textit{Update primal} variables.
        \FOR{$\text{batch } =1, \ldots, T_b \: $ } 
        \STATE 
        $
        \boldsymbol{\theta} \: \leftarrow \: \boldsymbol{\theta} - \eta_p \nabla_{\theta}\hat{L}(\boldsymbol{\theta}, \boldsymbol{\lambda}) 
        $
        \ENDFOR \\ \vskip 0.05in
    \hspace*{2mm} \textit{Evaluate constraint slacks.}
    \STATE
    $
    \mathbf{s}_l \quad \leftarrow \: \left( \frac{1}{N_b} \sum_{i=1}^{N_b} d_l(f_{\mathbf{\theta}_l}(\mathbf{z_i}^q_{l-1}), f^q(\mathbf{z_i}^q_{l-1})) \right) - \epsilon_{l}
    $
    \STATE
    $
    s_{\text{out}} \:  \: \leftarrow  \: \left( \frac{1}{N_b} \sum_{i=1}^{N_b}
    d(f_{\mathbf{\theta}}(\mathbf{x}_i), f^q(\mathbf{x}_i)) \right) - \epsilon_{\text{out}}
    $ \\ \vskip 0.05in
    \hspace*{2mm}\textit{ Update dual variables.}
    \STATE
    $
    \mathbf{\lambda}_{l} \quad \leftarrow \: \left[\lambda_{l} + \eta_d s_{l} \right]_{+}
    $
    
    \STATE
    $
    \lambda_{\text{out}} \: \:  \leftarrow \: \left[\lambda_{\text{out}} + \eta_d s_{\text{out}} \right]_{+}
    $
    \\
    \ENDFOR \\
    \STATE {\bfseries Return:}{  $\boldsymbol{\theta}$, $\boldsymbol{\lambda}$.} 
    \end{algorithmic}
\end{algorithm}

The advantage of~\eqref{ED-QRL} is that it is an unconstrained problem that, provided we have enough samples and the parametrization is rich enough, can approximate the constrained statistical problem~(\ref{QRL}). Namely, the difference between the optimal value of the empirical dual $D^{\star}_{\text{emp}}$ and the statistical primal $P^{\star}$, i.e., the empirical duality gap, is bounded (see \cite{LuizNonConvex}, Theorem 1). Observe that the function $\min _{\boldsymbol{\theta} \in \Theta} \hat{L}(\boldsymbol{\theta}, \boldsymbol{\lambda})$ is concave, since it corresponds to the minimum of a family of affine functions on $\boldsymbol{\lambda}$. Thus, the outer problem in \eqref{ED-QRL} is the maximization of a concave function and can be solved via gradient ascent. The inner minimization, however, is typically non-convex, but there is empirical evidence that over-parametrized neural networks can attain \textit{good} local minima when trained with stochastic gradient descent. The max-min problem~\eqref{ED-QRL} can be undertaken by alternating the minimization with respect to $\boldsymbol{\theta}$ and the maximization with respect to $\boldsymbol{\lambda}$  \cite{arrowhurwitz}, which leads to the primal-dual constrained learning procedure in Algorithm \ref{alg:pd}. 

In contrast to regularized objectives, where regularization weights are dimensionless quantities and typically require extensive tuning, constraint upper bounds can be set using domain knowledge. For instance, when comparing the absolute distance between high and low precision activations, it is reasonable to set $\epsilon_l$ in a neighbourhood of the minimum positive number representable in low-precision, i.e: $\frac{1}{2^n-1}$. 


\vspace{-0.1in}
\section{Experiments}\label{section:Exp}

\begin{figure*}[t]
    \centering
    \begin{subfigure}{0.33\textwidth}
    \centering
         \includegraphics[trim=0in 0.1in 0in 0in, clip,width=0.95\textwidth]{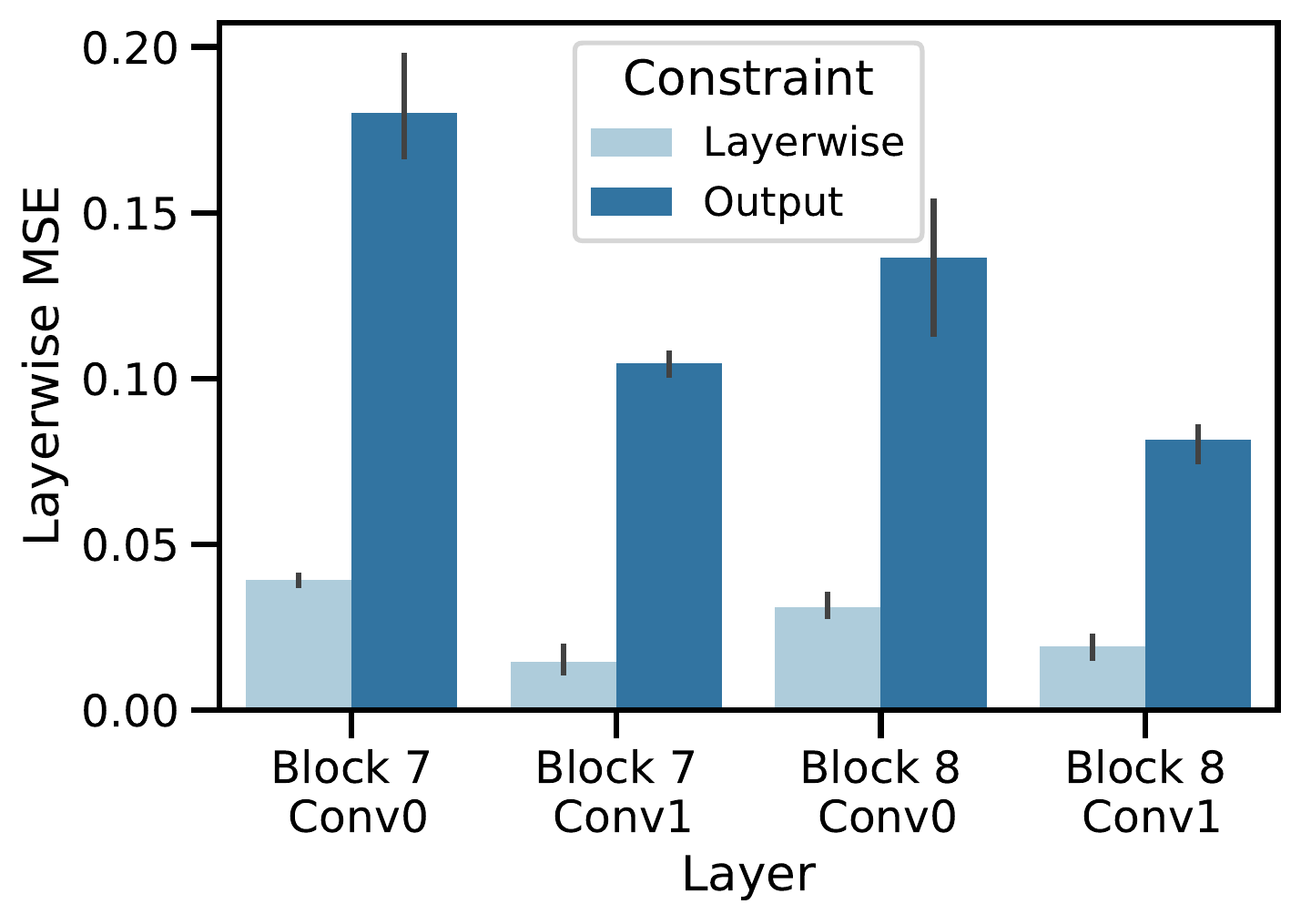}
     \caption{ }
     \label{lwa}
     \end{subfigure}
    \hfill
    \begin{subfigure}{0.33\textwidth}
    \centering
         \includegraphics[trim=0in 0.1in 0in 0in, clip,width=0.95\textwidth]{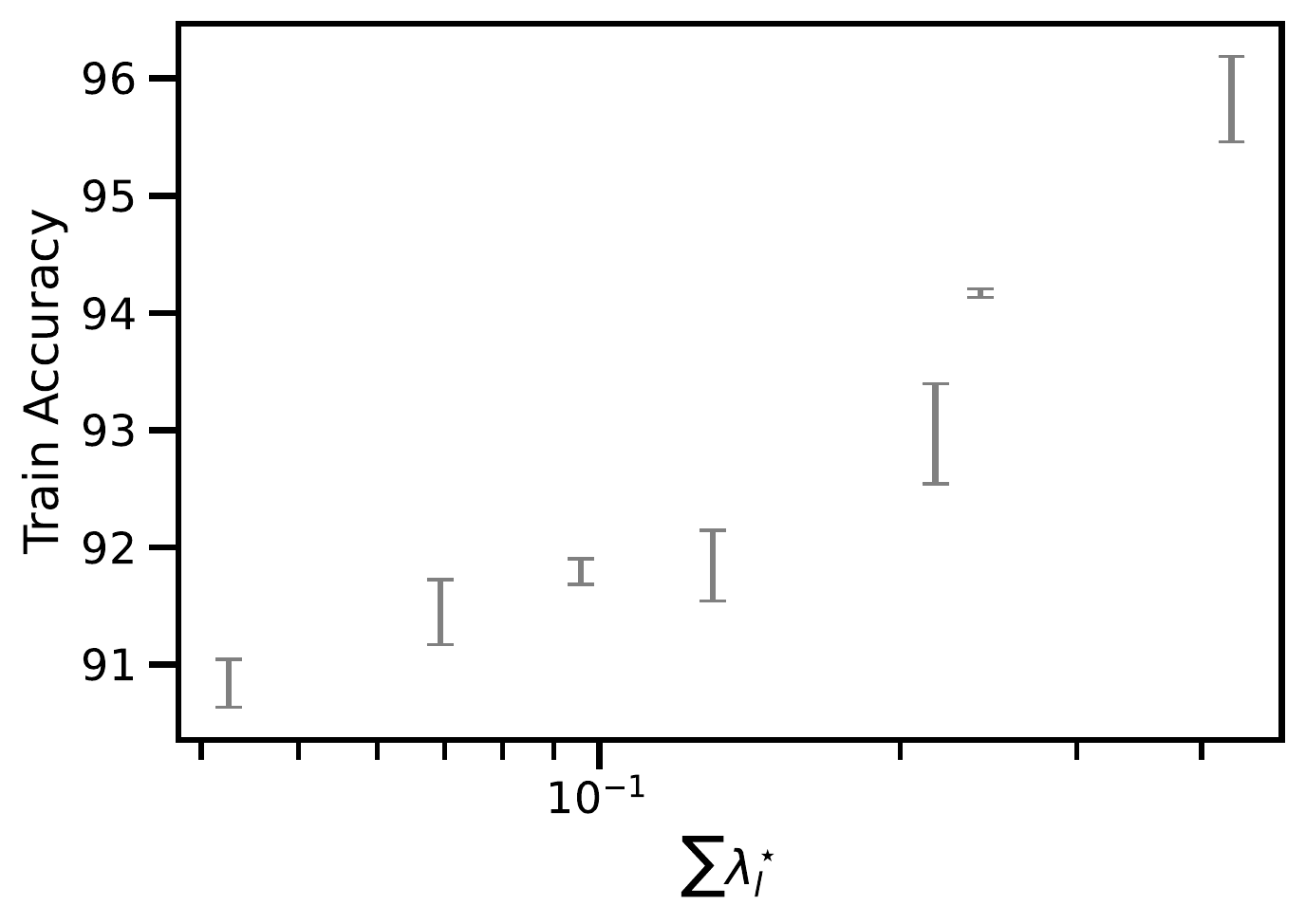}
        \caption{ }
        \label{lwb}
     \end{subfigure}
     \hfill
     \begin{subfigure}{0.33\textwidth}
        \centering
         \includegraphics[trim=0.03in 0in 0in 0.03in, clip, width=0.95\textwidth]{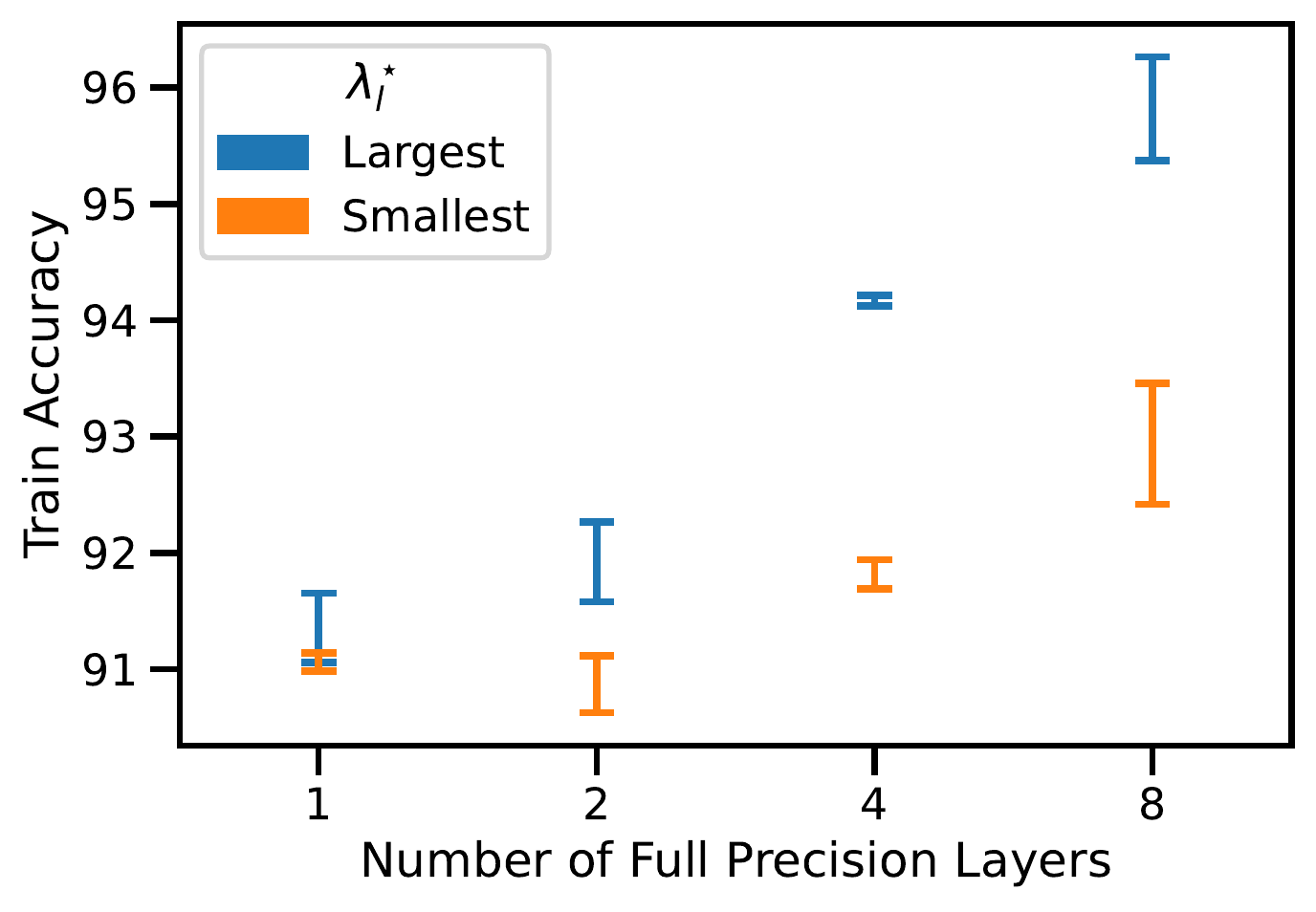}
     \caption{ }
     \label{lwc}
     \end{subfigure}
    \vskip -0.1in
    \caption{a) Mean-Squared-Error between the high and low precision activations at different layers, when using only output constraints and both output and layer-wise constraints.  b) Train accuracy with respect to the sum of the dual variables associated to the layers implemented in high precision. c) Train accuracy with respect to the number of layers implemented in high precision when selecting the layers with largest or smallest associated dual variables. Error bars denote the standard deviation computed across five runs.}
\end{figure*}

This section showcases Algorithm 1 in the CIFAR-10~\cite{CIFAR10} image classification benchmark using a ResNet-20~\cite{Resnet} architecture. First, we compare the performance of our constrained learning approach- with and without intermediate layer constraints- to baseline methods. Then, we show how Theorem 1 can be leveraged to select layers that, when implemented in high precision, have a large impact in model performance. We follow the experimental setup of~\cite{dorefa}. Additional experimental details can be found in Appendix~\ref{app:expdetails}.

In all experiments, optimization is run in high precision by applying simulated quantization to weights and activations as in the usual QAT setting~\cite{whitepaper-quant}. We use the popular quantization scheme proposed by~\cite{dorefa}, which has been shown to achieve comparable performance to state of the art methods (e.g.~\cite{LSQ}) in standarised benchmark settings~\cite{MQBench}. Namely, we quantize the weights $\mathbf{w}$ and activations $\mathbf{a}$ at each layer according to
\vskip -0.21in
\begin{align*}
    &q_w(\mathbf{w}) = 2 \: r\left(\frac{1}{2}+\frac{\text{tanh}(\mathbf{w})}{2\max(|\text{tanh}(\mathbf{w})|)}\right)-1,\\
    &q_a(\mathbf{a}) = r\left( \text{clip}_{0\leq a \leq 1}(\mathbf{a}) \right),
\end{align*} where $r(z) = \frac{1}{2^k-1}\text{round}(z\times(2^k-1))$ maps $z\in\mathbb{R}$ to its $k$-bit fixed point representation.

To enforce proximity at the model's output and at intermediate activations we use cross-entropy distance and mean-squared error, respectively. That is, 
\begin{align*}
  & d_l(f_l(\mathbf{z}^q_{l-1}), f_l^q(\mathbf{z}^q_{l-1})) = \|f_l(\mathbf{z}^q_{l-1})-f_l^q(\mathbf{z}^q_{l-1}) \|^2_2 \\
  & d(f(x), f^q(x)) = -\sum_i f(\mathbf{x})_i \log(f^q(\mathbf{x})_i).
\end{align*}

\vskip -0.1in
As shown in Table \ref{table}, both variants of PDQAT exhibit competitive performance, outperforming the baseline in all precision levels by a small margin. For all methods, $8$ and $4$ bit quantizations do not hinder test accuracy significantly. Lower precision levels manifest a larger gap (e.g: ~6.5 \% at $1$ bit), the drop being less severe for our primal dual approach.

Requiring proximity at intermediate layers impacts the properties of the low precision models in various ways. At most precision levels, it induced a slight increase in test performance (see Table \ref{table}). Moreover, as shown in Figure \ref{lwa}, the distance between activations at intermediate layers is approximately an order of magnitude smaller than those of a model trained without layer-wise constraints. Finally, it enables the use of dual variables as indicators of the potential benefit of implementing each individual layer in high precision, as shown in Figures \ref{lwb} and \ref{lwc}. For instance, as the sum of the dual variables associated to the layers implemented in high precision increases, so does the accuracy of the mixed precision model. Similarly, implementing layers with large associated dual variables in high precision is more impactful than implementing those with small associated dual variables.
\vspace{-0.12in}
\section{Related Work}\label{section:Related}
\vskip -0.1in
Several works \cite{kd-2018, kd-apprentice-2018, KD-2020, kd-fisher-2020, Kd-AnyPrec-2021} have proposed to use Knowledge Distillation techniques for quantization aware training. These optimise a regularised objective comprised of the quantised model's loss and a term that promotes proximity between its outputs and the outputs of a full precision \emph{teacher} model. Unlike our approach, these methods rely on STE to backpropagate through the quantization operation. Furthermore, whereas these two terms are weighted by a fixed penalisation coefficient, our primal dual approach dynamically adjusts strength of the proximity term during training.

\cite{HAWQV2,HAWQV3} propose to estimate the sensitivity of a layer to quantization using the Hessian of pre-trained models. Therefore, these methods only contemplate the final model, whereas our approach accounts for training dynamics. Other proposed layer selection techniques, such as reinforcement learning ~\cite{RL-quant} and differentiable neural architecture search~\cite{DNAS-quant}, are computationally intensive due to the large search space.

\begin{table}
\begin{tabular}{c|c|cc}
\toprule
 & & \multicolumn{2}{c}{OURS}\\
 Bitwidth & Baseline & (i) & (ii)  \\ 
\hline  32 & \multicolumn{3}{c}{$91.50 \pm 0.26 $}\\
\hline  8  & $ 91.34 \pm 0.24 $  & $ 91.51 \pm 0.27 $& $91.47 \pm 0.13 $ \\
 4  & $ 91.38 \pm 0.16 $  & $ 91.20 \pm 0.14 $& $91.48 \pm 0.23 $ \\
 2  & $ 89.38 \pm 0.34 $  & $ 89.43 \pm 0.24 $& $89.63 \pm 0.14 $ \\
 1  & $ 84.91 \pm 0.36 $  & $ 85.29 \pm 0.27 $& $85.52 \pm 0.33 $ \\
\hline  
 \end{tabular}
\caption{Performance comparison of our method with respect to DoReFa~\cite{dorefa}. (i) Denotes PDQAT with a proximity constraint only at the models' outputs and (ii) denotes PDQAT with proximity requirements both at the output and intermediate layers. We report the mean and standard deviation across five runs.}
\label{table}
\end{table}

\vspace{-0.12in}
\section{Conclusion}\label{sec:conclu}

In this paper, we presented a constrained learning approach to quantization aware training. We showed that formulating low precision supervised learning as a strongly dual constrained optimization problem does away with gradient estimations. We also demonstrated that the proposed approach exhibits competitive performance in image classification tasks. Furthermore, we showed that dual variables indicate the sensitivity of the objective with respect to constraint perturbations. We leveraged this result to apply layer selective quantization based on the value of dual variables, leading to considerable performance improvements. Analyzing the performance of the proposed approach in more realistic scenarios, and imposing constraints to intermediate outputs at other granularity levels, e.g. in a channelwise fashion, are promising future work directions.
\blfootnote{The authors thank Alejandro García (UdelaR) for meaningful discussions.}

\clearpage
\bibliographystyle{IEEEbib}
\bibliography{references}

\appendix

\section{Quantized Convolutional Block}
\begin{figure}[h]
    \centering
    \includegraphics[width=0.3\textwidth]{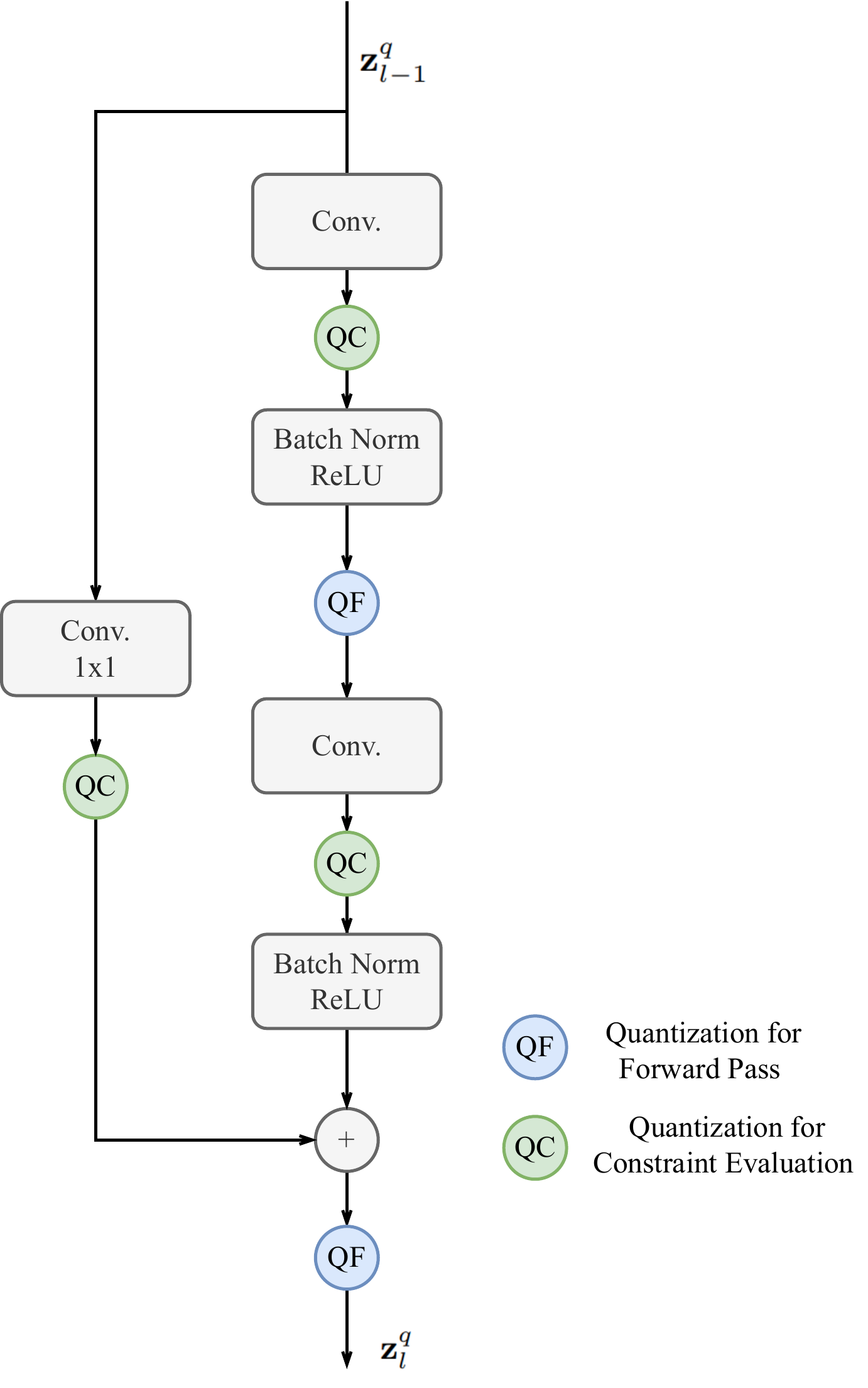}
    \caption{Diagram of low precision residual block.}
    \label{fig:diag}
\end{figure}

\section{Dynamics of Dual Variables}

\begin{figure}[h]
    \centering
    \includegraphics[width=0.5\textwidth]{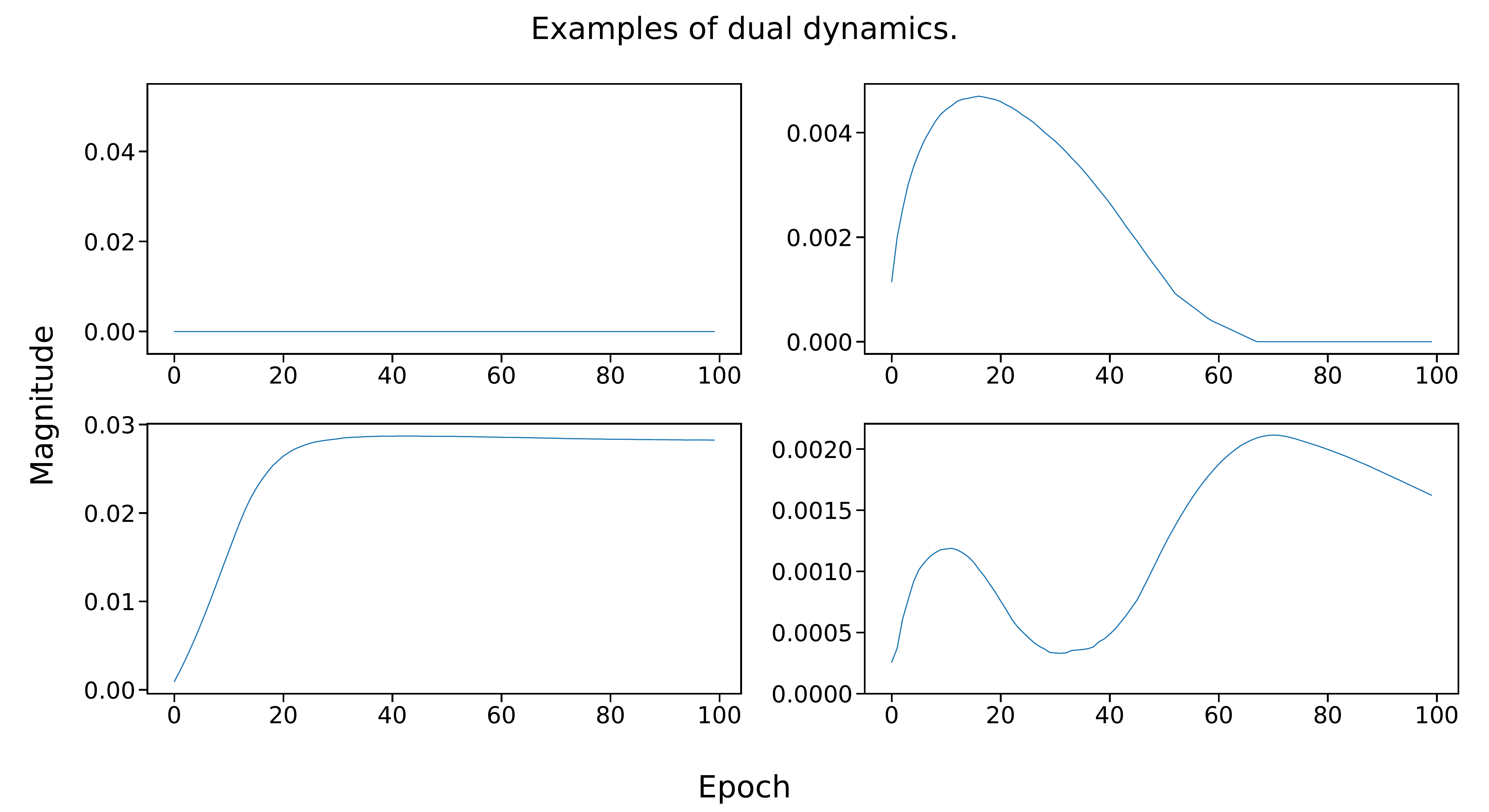}
    \caption{Evolution of dual variables during training a 2-bit model with L2 layer-wise constraints and cross-entropy constraint on the output.}
    \label{fig:dualdynamics}
\end{figure} 

As explained in section \ref{sec:prob-form}, the final dual variables indicate the sensitivity of the optimal value with respect to perturbations in the constraints. However, as shown in the top row of Figure \ref{fig:dualdynamics}, dual variables that are zero at the end of the training procedure may exhibit different dynamics. Thus, their respective constraints have different impact on the objective throughout training. \\
In contrast, the bottom row shows dual variables that are active throughout the training procedure, but their evolution is different. The dual variable shown on the bottom-left figure increases until the constraint is satisified and then remains constant (its slack becomes zero), while the one on the bottom-right figure exhibits an increasing oscillation.

\section{Proof of Theorem 1 (Sensitivity of $P^{\star}_c$)}
\label{app:sensi}
This result stems from a sensitivity analysis on the constraint of problem \eqref{QRL} and is well-known in the convex optimization literature. More general versions of this theorem are shown in \cite{shapirobook} (Section 4).
Let $\mathbf{\lambda} = (\lambda_1, \cdots, \lambda_L, \lambda_{\text{out}})$. We start by viewing $P^{\star}_c$ as a function of $\epsilon_l$:
\begin{align*}
P^{\star}_c(\epsilon_l) =  & \min_{f \in \mathcal{F}}  \quad  \; \mathbf{E}_{(x, y)\sim \mathfrak{D}} \left[\ell \left(f(\x), y \right)\right], \\
& \text{s.t. :} \quad   \mathbb{E}_{\mathfrak{D}} \left[ d_i(f_i(\mathbf{z}^q_{i-1}), f_i^q(\mathbf{z}^q_{i-1})) \right] \leq \epsilon_i  \quad   i = 1,\cdots,L-1 \nonumber\\
  & \quad \quad \: \: \: \: \mathbb{E}_{\mathfrak{D}}\left[ d(f(x), f^q(x)) \right] \leq \epsilon_{\text{out}} 
\end{align*}
The Lagrangian $L(f, \lambda; \: \epsilon_l)$ associated to this problem can be written as 
\begin{align*}
 L(f, \lambda; & \epsilon_l) = \mathbf{E}_{(x, y)\sim \mathfrak{D}}  [ \: \: \ell \left(f(\x), y \right) \\ & +  \sum_{i=1, i \neq l }^{L-1} \lambda_i \left( d_i(f_i(\mathbf{z}^q_{i-1}), f_i^q(\mathbf{z}^q_{i-1})) - \epsilon_i \right)  \\ & + \lambda_l \left(d_l(f_l(\mathbf{z}^q_{l-1}), f_l^q(\mathbf{z}^q_{l-1})) - \epsilon_l\right) \\ & + 
\lambda_{out} \left( d(f(x), f^q(x)) -  \epsilon_{\text{out}} \right) \: \: ]
\end{align*}
where the dependence on $\epsilon_l$ is explicitly shown. Then, following the definition of $P^\star_c(\epsilon_l)$ and using strong duality, we have $$P^\star_c(\epsilon_l)=\min_f L(f, \lambda^{\star}(\epsilon_l); \epsilon_l) \leq L(f, \lambda^{\star}(\epsilon_l); \epsilon_l)$$ with the inequality being true for any function $f \in \mathcal{F}$, and where the dependence of $\lambda^{\star}$ on $\epsilon_l$ is also explicitly shown. Now, consider an arbitrary function $\epsilon' \in \mathbb{R^+}$ and the respective primal function $f^{\star}(\cdot;\epsilon')$ which minimizes its corresponding Lagrangian. Plugging $f^{\star}(\cdot;\epsilon')$ into the above inequality, we have
\begin{align*}
    P^\star_c(\epsilon_l) &\leq L(f^{\star}(\cdot;\epsilon'), \lambda^{\star}(\epsilon_l); \epsilon_l) \\ &= \mathbf{E}_{(x, y)\sim \mathfrak{D}}  [ \: \ell \left(f^{\star}(\mathbf{x};\epsilon'), y \right) \\ & +  \sum_{i=1, i \neq l}^{L-1} \lambda^{\star}_i(\epsilon_l) \left( d_i(f^{\star}_i(\mathbf{z}^q_{i-1};\epsilon'), {f^{\star}}^q_i(\mathbf{z}^q_{i-1};\epsilon')) - \epsilon_i \right) \\ 
    & + \lambda^{\star}_l(\epsilon_l) \left( d_l(f^{\star}_l(\mathbf{z}^q_{l-1};\epsilon'), {f^{\star}}^q_l(\mathbf{z}^q_{l-1};\epsilon')) - \epsilon_l \right)
    \\ & +
    \lambda^{\star}_{out}(\epsilon_l) \left( d(f^{\star}(\mathbf{x};\epsilon'), {f^{\star}}^q(\mathbf{x};\epsilon') ) -  \epsilon_{\text{out}} \right) \: ]
\end{align*}
Now, since $f^{\star}(\cdot;\epsilon')$ is \emph{optimal} for constraint bounds given by $\epsilon'$ and complementary slackness holds, we have
$$\mathbb{E}_{(\boldsymbol{x}, y) \sim \mathfrak{D}} \: \Big[ \ell (f^{\star}(\boldsymbol{x}; \epsilon'), y)  \Big] = P^\star_c(\epsilon').$$ Moreover, $f^{\star}(\cdot;\epsilon')$ is, by definition, feasible for constraint bounds given by $\epsilon_1, \cdots, \epsilon', \cdots,\epsilon_L,$ and $\epsilon_{\text{out}}$. In particular,
\begin{align*}
& d_l(f^{\star}_l(\mathbf{z}^q_{l-1};\epsilon'), {f^{\star}}^q_l(\mathbf{z}^q_{l-1};\epsilon')) \leq \epsilon'
\end{align*}
This implies that,
\begin{align*}
 & d_l(f^{\star}_l(\mathbf{z}^q_{l-1};\epsilon'), {f^{\star}}^q_l(\mathbf{z}^q_{l-1};\epsilon')) - \epsilon_l \\
&= d_l(f^{\star}_l(\mathbf{z}^q_{l-1};\epsilon'), {f^{\star}}^q_l(\mathbf{z}^q_{l-1};\epsilon')) - \epsilon' + \epsilon' - \epsilon_l \\
&= \alpha + (\epsilon' - \epsilon_l) \quad \text{with } \: \alpha \leq 0
\end{align*}
Combining the above, we get
$$
P^\star_c(\epsilon_l) \leq P^\star_c(\epsilon') + \lambda^{\star}_l(\epsilon_l)( \epsilon'  - \epsilon_l ) 
$$ or eqivalently,
\begin{align*}
 P^\star_c(\epsilon')  \geq P^\star_c(\epsilon) -\lambda^{\star}(\epsilon_l)( \epsilon'  - \epsilon_l ),
\end{align*} which matches the definition of the subdifferential, hence completing the proof. The proof for $\epsilon_{\text{out}}$ follows the same steps, considering index $L$ in the dual variable vector.

\section{Ablation on Constraint Tightness}

\subsection{Output constraint}

\begin{table}[h]
\centering
\begin{tabular}{|c|cccc|}
\hline
Bitwidth           & \multicolumn{4}{c|}{}                                                                                \\ \hline
\multirow{2}{*}{4} & \multicolumn{1}{c|}{$\epsilon_{\text{out}}$}   & \multicolumn{1}{c|}{0.1}   & \multicolumn{1}{c|}{0.2}   & 0.7   \\ \cline{2-5} 
                   & \multicolumn{1}{c|}{Test Accuracy} & \multicolumn{1}{c|}{91.39} & \multicolumn{1}{c|}{91.33} & 91.27 \\ \hline
\multirow{2}{*}{2} & \multicolumn{1}{c|}{$\epsilon_{\text{out}}$}   & \multicolumn{1}{c|}{0.05}  & \multicolumn{1}{c|}{0.2}   & 0.7   \\ \cline{2-5} 
                   & \multicolumn{1}{c|}{Test Accuracy} & \multicolumn{1}{c|}{89.00} & \multicolumn{1}{c|}{89.44} & 89.50 \\ \hline
\multirow{2}{*}{1} & \multicolumn{1}{c|}{$\epsilon_{\text{out}}$}   & \multicolumn{1}{c|}{0.5}   & \multicolumn{1}{c|}{0.8}   & 1.5   \\ \cline{2-5} 
                   & \multicolumn{1}{c|}{Test Accuracy} & \multicolumn{1}{c|}{84.08} & \multicolumn{1}{c|}{85.33} & 85.37 \\ \hline
\end{tabular}
\caption{Ablation on output constraint tightness in CIFAR-10 with ResNet-20.}
\end{table}

\subsection{Layerwise constraint}

\begin{table}[h]
\centering
\begin{tabular}{|c|cccc|}
\hline
Bitwidth           & \multicolumn{4}{c|}{}                                                                                 \\ \hline
\multirow{2}{*}{4} & \multicolumn{1}{c|}{$\epsilon_l$}       & \multicolumn{1}{c|}{0.082}  & \multicolumn{1}{c|}{0.16}  & 0.25  \\ \cline{2-5} 
                   & \multicolumn{1}{c|}{Test Accuracy} & \multicolumn{1}{c|}{91.39}  & \multicolumn{1}{c|}{91.38} & 91,48 \\ \hline
\multirow{2}{*}{2} & \multicolumn{1}{c|}{$\epsilon_l$}       & \multicolumn{1}{c|}{0.055}  & \multicolumn{1}{c|}{0.25}  & 0.67  \\ \cline{2-5} 
                   & \multicolumn{1}{c|}{Test Accuracy} & \multicolumn{1}{c|}{89.54}  & \multicolumn{1}{c|}{89.58} & 89.67 \\ \hline
\multirow{2}{*}{1} & \multicolumn{1}{c|}{$\epsilon_l$}       & \multicolumn{1}{c|}{0.0125} & \multicolumn{1}{c|}{0.25}  & 1     \\ \cline{2-5} 
                   & \multicolumn{1}{c|}{Test Accuracy} & \multicolumn{1}{c|}{83.5}   & \multicolumn{1}{c|}{85.19} & 85.40 \\ \hline
\end{tabular}
\caption{Ablation on layer-wise constraint tightness in CIFAR-10 with ResNet-20.}
\end{table}

\section{Additional Experimental Details}\label{app:expdetails}

The code for reproducing all experiments in this paper is available at \url{http://github.com/ihounie/pd-qat}.

\subsection{Simulated quantization}

As in related works~\cite{dorefa} we perform simulated quantization, that is, we quantize weights and activations but perform operations in full precision. Figure~\ref{fig:diag} explicits where quantization operations are applied during forward passes and constraint evaluation for a Resnet~\cite{Resnet} block. Following~\cite{dorefa, BengioBinarizedNN,XNORNetIC} and other related works, we do not quantize the input and output layers. Unlike~\cite{Kd-AnyPrec-2021} we quantize all other convolutional layers, including shortcuts.

\subsection{Batchnorm Layers}
The full and low precision models have different activation statistics, which can hinder training and performance due to batch normalisation layers~\cite{oscilations-qat}. In order to overcome this, we simply use different batch normalisation layers for each precision level. As in related works~\cite{dorefa} we do not perform batch normalisation folding.

\subsection{Hyperparameters}
We train models for a maximum of 100 epochs using Adam~\cite{Adam} without weight decay, initial learning rate 0.001
decayed by 0.1 at epochs 50, 75 and 90, and perform early stopping with validation accuracy as a stopping criteria. We use a dual learning rate of 0.01.

\end{document}